\documentclass{article}

\PassOptionsToPackage{numbers, compress}{natbib}

\usepackage[preprint]{neurips_2023}

\title{Several questions of visual generation in 2024}

\author{
    Shuyang Gu \\
    Microsoft Research Asia \\
    \texttt{cientgu@gmail.com} \\
}

\usepackage{times}
\usepackage[utf8]{inputenc} %
\usepackage[T1]{fontenc}    %
\usepackage{hyperref}       %
\usepackage[dvipsnames]{xcolor}
\hypersetup{
  colorlinks,
  linkcolor={red!50!black},
  citecolor={blue!50!black},
  urlcolor={blue!80!black}
  }
  \definecolor{orange}{HTML}{ff7f0e}
  \definecolor{blue}{HTML}{1f77b4}
  
\usepackage[framemethod=TikZ]{mdframed}
\usepackage{url}            %
\usepackage{booktabs}       %
\usepackage{amsfonts}       %
\usepackage{nicefrac}       %
\usepackage{microtype}      %
\usepackage{comment}     
\usepackage{amssymb}
\usepackage{amsmath,amsfonts,bm}
\usepackage{nicefrac}
\usepackage{textcomp}
\usepackage[capitalise]{cleveref}
\usepackage{enumitem}%
\usepackage{subcaption}
\usepackage{floatrow}
\usepackage{tabularx}
\usepackage{makecell}
\usepackage{mathtools}
\usepackage[most]{tcolorbox}
\usepackage{nicefrac}
\usepackage{booktabs}
\usepackage[]{mdframed}
\usepackage{wrapfig}
\usepackage{arydshln}
\usepackage{colortbl}

\usepackage{empheq}
\usepackage[most]{tcolorbox}

\newtheorem{definition}{Definition}
\DeclareMathOperator*{\argmin}{arg\,min}

\begin{document}

\maketitle

\begin{abstract}
  This paper does not propose any new algorithms but instead outlines various problems in the field of visual generation based on the author's personal understanding. The core of these problems lies in \emph{how to decompose visual signals}, with all other issues being closely related to this central problem and stemming from unsuitable approaches to signal decomposition. This paper aims to draw researchers' attention to the significance of \textbf{Visual Signal Decomposition}.
  
  Due to the limitations of the author's knowledge, the paper does not cover certain critical issues, such as security issues with generative models and data annotation challenges. Moreover, it is possible that many of the opinions in this paper are prejudiced or even incorrect. I sincerely welcome readers to refute them or to discuss with me.

\end{abstract}

\section*{Question 1: What's the goal of generative models?}
In my view, generative models are designed to "create what one envisions". The generation of digital signals requires three milestones: 

1. The first milestone involves translating the user's thoughts into a format interpretable by a computer~\cite{shneiderman1982future,wolpaw2002brain,hochberg2012reach}. Specifically, this entails identifying the modality of the signal to be generated, which could range from text, 3D objects, to videos, or other forms. Subsequently, one must ascertain the exact state that is to be generated under the given conditions. From the perspective of manifold learning, this process corresponds to first determining the dimensionality $D$ of the encompassing space, followed by identifying the target distribution $P_t$ that the model seeks to emulate.

2. In the field of generative modeling, numerous researchers are engrossed in the challenge of how to modeling $P_t$. The principal obstacle is the inherent complexity of the target distribution, which often makes it exceedingly arduous to model. Consequently, researchers are continuously seeking out models with stronger modeling capabilities. Specifically, within the realm of visual generation, prevalent methodologies in recent years include Energy-Based Models~\cite{lecun2006tutorial}, Variational Autoencoders (VAEs)~\cite{kingma2013auto}, Generative Adversarial Networks (GANs)~\cite{goodfellow2014generative}, diffusion models~\cite{ho2020denoising,song2020score}, and many others.

3. Simultaneously, many researchers further exploring more advanced objectives, investigating how to generate results with greater efficiency~\cite{song2023consistency, meng2023distillation} and enhanced interpretability~\cite{tang2022daam, geng2024instructdiffusion}. These pursuits have substantial implications for practical deployment, security and the development of responsible AI.

\section*{Question 2: The problem of visual signal decomposition.}
As discussed in Question 1, the endeavor to model the distribution $P_t$ presents a significant challenge, especially when the goal is to generate convincing textual content or videos. These types of data represent low-dimensional manifolds embedded within extremely high-dimensional spaces, making the direct fitting of these distributions with a single network impractical. Therefore, we need to break down the complex distribution modeling problem into multiple simpler problems and solve each subproblem separately. This gives rise to a question: \emph{How to effectively decompose this complex distribution modeling problem?}

\subsection*{Language decomposition}
\emph{The success of large language models is largely attributed to the effective decomposition of textual signals}. Consider the task of modeling a text sequence $A={x_0, x_1, x_2, \ldots }$, where $x_t$ denotes the token at the $t$-th position from front to back. This can be segmented into a sequence of conditional modeling subtasks based on the position: $p(x_0), p(x_1|x_0), p(x_2|x_1, x_0), \ldots $. Large language models employ autoregressive models to approximate these conditional distribution mappings $p(x_t|x_{i, i<t})$. A critical aspect of this approach is that, for natural language, the decomposed subtasks are intrinsically interrelated. For example, a phrase is "I love swimming," it might appear at the beginning or in the middle of a sentence. In other words, it can appear in any subtask. With a sufficiently large dataset, $p(x_2="\text{swimming}"|x_{i<2}="\text{I love}")$ and $p(x_7="\text{swimming}"|x_{i<7}="\text{you love playing basketball, I love}")$  represent two tasks that are very interrelated and can serve as a form of data augmentation to one another. In other words, different subtasks are \textbf{"equivariant"}. Let's give "equivariant" a rigorous mathematical definition.

\begin{definition}
     Assume the target distribution is $P_t(x)$, we split the signal into multiple subtasks: $p(x_0), p(x_1|x_0), p(x_2|x_1) \ldots p(x_{t+1}|x_t) \ldots$ For the $t$-th conditional probability fitting task $p(x_{t+1}|x_t)$, we adopt a network $\theta_t$ to fit it. For any two tasks $t$ and $k$ and two state samples $s$ and $\hat{s}$, if 
    \begin{equation}
     p_{\theta_t}(x_{t+1}=\hat{s}|x_{t}=s) = p_{\theta_k}(x_{t+1}=\hat{s}|x_{t}=s)   
     \label{eq:equivariant_1}
    \end{equation}
    or
    \begin{equation}
     p_{\theta_t}(x_{t+1}=\hat{s}|x_{t}=s) = p_{\theta_t}(x_{k+1}=\hat{s}|x_{k}=s)   
     \label{eq:equivariant_2}
    \end{equation}
    we refer to this signal decomposition as equivariant.
\end{definition}

It can be observed that the language decomposition is independent of position. For any token $x_t$ or phrase $\{x_t,x_{t+1}\}$, the probability of their occurrence at the t-th or k-th position in a sentence is nearly identical.\footnote{Although the word "I" may have a higher probability of appearing in the first position than others, given the typically extensive length of linguistic texts, it can be approximated as identical.} This observation aligns with Equation~\ref{eq:equivariant_2}, thus suggesting that the language decomposition exhibits equivariance. Therefore, employing a single model to approximate these diverse but related tasks generally does not lead to conflicts, and in fact, it can be highly beneficial for the overall modeling of the data.

\subsection*{Image patch decomposition}
\begin{figure}
    \centering
    \includegraphics[width=1\textwidth]{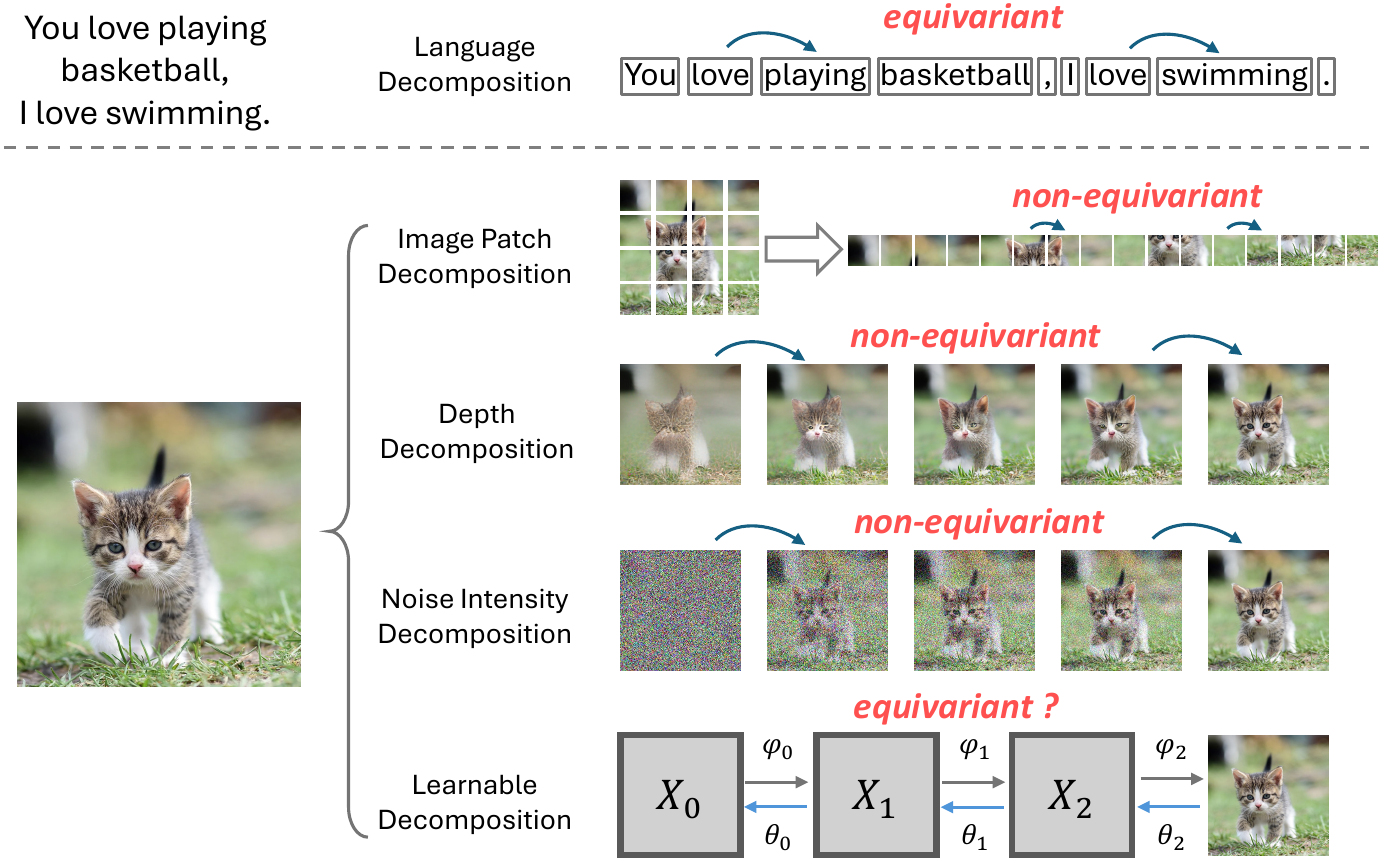}
    \caption{
        Illustration of Language and Visual Signal Decomposition. The language decomposition is equivariant, while most image decomposition are non-equivariant.
    }
    \label{fig:SignalDecomposition}
\end{figure}

This strategy of decomposition and modeling has made remarkable success in language domains. However, replicating this approach to decompose image through spatial position presents distinct challenges. Early attempts~\cite{chen2020generative,ramesh2021zero} involved segmenting images into spatial patches to create sequences that could be processed by autoregressive models. Yet, unlike natural language, image patch inherently lack the "equivariance" characteristic. As illustrated in Figure~\ref{fig:SignalDecomposition}, while there is a continuity within the blocks of a single row, such continuity is absent between the last block of one row and the first block of the subsequent row. In addition to continuity, other dataset-specific challenges persist, such as the propensity for human subjects to be centered within images. Therefore, employing a single, universal model to encapsulate all these varying distributions often results in conflicts. Moreover, the model must endeavor to learn an intricate joint distribution mapping problem that composed of multiple different distributions, which undermines the principle of decomposing complex distributions for simpler modeling. Although the integration of position-embedding can mitigate some of these conflicts, it is not a panacea for the issue.

\subsection*{Depth decomposition}
Beyond spatial division, some researchers have explored splitting image data along the depth dimension. This approach is intuitively appealing, given the large amount of information present at each spatial location in an image, such as the need for a three-channel RGB representation with 8 bits per channel. On the other hand, the high spatial dimensionality of images often necessitates the use of Variational Autoencoders (VAE)~\cite{kingma2013auto} to reduce the dimension. Combining these two aspects, notable methods that partition data by depth include VQVAE2~\cite{razavi2019generating} and RQVAE~\cite{lee2022autoregressive}. Firstly, such approaches may encounter the issue of "invalid encoding," which we will elaborate in Question3. Furthermore, unlike natural language, the depth dimension also does not exhibit "equivariance". Taking RQVAE as an example, at the same spatial position, earlier tokens represent low-frequency information while later tokens represent high-frequency information. Therefore, applying autoregressive models with shared parameters~\cite{lee2022autoregressive, tian2024visual} to model these varied distribution mappings can result in conflicts. An alternative approach, such as MUSE~\cite{chang2023muse}, proposes segmenting the problem into a smaller number of unique subtasks, each modeled independently without parameter sharing. However, as the data distributions become increasingly complex, this may necessitate a larger set of subtasks, leading to a surge in the required model parameters and potentially exacerbating the "invalid encoding" problem.

\subsection*{Noise intensity decomposition}
Diffusion models propose another intriguing approach to signal decomposition: characterizing images through a series of progressively denoised image sequences. 
For a given image $x_0$ in a dataset, noise is sequentially added via a Markov process to produce a sequence ${x_0, x_1, x_2, \ldots, x_N}$, with $x_N$ being nearly pure noise, with minimal remnants of the original image information. This process effectively breaks down the task of modeling the image distribution into N denoising subtasks: $p(x_t|x_{t+1})$, for $t=[0,1,\ldots,N-1]$. While all subtasks are concerned with denoising, parameter sharing among them might seem practical in theory. However, in practice, typical noise-adding strategies can lead to discrepancies during the denoising stages, especially when noise levels differ significantly~\cite{hang2023efficient}. These non-equivariant tasks lead to a predicament akin to depth-based decomposition challenge: employing a model with shared parameters to fit complex data distribution mappings is a daunting task for the model's capability. If parameters are not shared, this could inflate the model size rapidly. Some studies, such as eDiff-I~\cite{balaji2022ediff}, have attempted to balance the parameter efficiency with the complexity of fitting distributions from an implementation standpoint. Additionally, the reparameterization trick~\cite{nichol2021improved, salimans2022progressive, lipman2022flow} has proven to be an extremely important technique to unify the output distributions of different denoising task, which alleviates the conflicts across different noise intensities. However, it does not eliminate the discrepancy of the input distributions. These concerns regarding noise intensity conflicts within diffusion models will be further explored in Question 4.

\subsection*{Learnable decomposition}
Upon reviewing the diffusion models, we find that the extent of conflicts is determined by the chosen noise strategy, which is typically predefined manually. Therefore, some researchers have attempted to define superior noise strategies, endeavoring to ensure a certain level of similarity in the denoising process across varying noise levels. Notable works in this area include Flow Matching~\cite{lipman2022flow} and consistency models~\cite{song2023consistency}. Concurrently, other researchers are exploring whether the adding noise strategy can be learned rather than preset. Noteworthy advancements in this domain comprise Variational Diffusion Models (VDM)~\cite{kingma2021variational} and Diffusion Schrödinger Bridge (DSB)~\cite{de2021diffusion,tang2024simplified}, although not all such works were initiated pursued this objective. In particular, VDM focuses on learning the coefficients for adding Gaussian noise, which somewhat narrows the potential for learning to mitigate conflicts. Meanwhile, studies based on the Schrödinger Bridge paradigm employ specialized networks to learn the adding noise process, iteratively approximating the entropy-regularized optimal transport. However, current learnable decomposition methods are not designed according to the characteristic of "equivariance". Future work may need to consider this as a prior, to constrain the network learning in the learnable decomposition. Moreover, a significant trade-off arises when substituting a predefined SDE with a network for noise learning: it's challenging to leverage the reparameterization to unify output distributions, which is a pivotal technique to mitigate conflicts across different noise intensities in practical applications. Although recent efforts~\cite{shi2024diffusion,tang2024simplified} have been made to address these issues, they still remain insufficient for practical application.

\subsubsection*{Extended Discussion}
From the signal decomposition perspective, the debate over whether autoregressive (AR) models, diffusion models, or other model architectures are superior for visual generation is not particularly productive. The essential consideration is how the signal is decomposed and the selection of a generative paradigm whose inductive bias is conducive to the chosen decomposition approach. Ideally, there may be two paradigms for decomposition: one resembles the approach taken by language models, which simplifies complex data distributions into a series of simpler, conditional data distributions that exhibit "equivariance". The other strategy involves breaking down the data into multiple independent distribution problems, which can be viewed as a special instance of "equivariance".

We argue that the difficulty of achieving equivariance in image decomposition is not entirely due to the fact that images are two-dimensional data while language is one-dimensional. Recent work~\cite{yu2024image} attempts to encode images to one-dimensional tokens, but these one-dimensional tokens are neither independent nor equivariant.

While learnable decomposition methods theoretically hold the potential of achieving this "equivariance", their practical application is fraught with challenges currently. Another feasible approach might be to integrate various signal decomposition techniques to simplify the data distribution. For instance, videos can be decomposed into temporally "equivariant" frames, which can then be further broken down based on "noise intensity"~\cite{blattmann2023stable} or "image patches"~\cite{kondratyuk2023videopoet}. Similarly, MUSE~\cite{chang2023muse} initially decomposes the image signal along the depth dimension and then tackles the distribution mapping from the "noise intensity" dimension. 

We argue the signal decomposition is the fundamental problem, and many of the subsequent questions can be seen as extensions of it, aims to alleviate the issue of non-equivariance in the current visual signal decomposition.

\section*{Question 3: The tokenization problem.}
Current mainstream generative models for images and videos predominantly adopt a two-stage approach: initially encoding the data into a compact, low-dimensional representation, then modeling this compressed distribution. The purpose of the first compression phase is to streamline the data distribution while preserving as much of the original information as possible. This simplification is intended to alleviate the complexity faced during the model fitting phase that follows. In the context of textual data, the dimensionality reduction can be considered lossless. In contrast, visual data compression, whether through Autoencoders (AE) or Variational Autoencoders (VAE), is inherently lossy. However, the assertion that "the less lossy the compression, the better" is not necessarily valid. A typical example is that signals compressed with AE might reconstruct better than those with VAE, but the complexity retained in the compressed data distribution still poses challenges for the fitting process in the second stage. Therefore, researchers introduce regularization constraints~\cite{van2017neural,rombach2022high,gu2022vector} in the latent space during compression to prevent the data distribution from becoming overly complex.

The conflict between reconstruction fidelity and fitting difficulty is a familiar challenge across various fields. In audio processing, continuous audio signals are typically encoded into tokens of length 16, with the fitting phase often concentrating on the foremost 8 tokens~\cite{wang2023neural}. Similarly, in the realm of image processing, GLOW~\cite{kingma2018glow} and VDM++~\cite{kingma2024understanding} also discovered that training on 5-bit images yields better results than using full 8-bit depth. These observations underscore the potential benefits of employing an adaptive-length encoding strategy to better balance the precision of reconstruction against the complexity of the second stage fitting task.

\begin{figure}
    \centering
    \includegraphics[width=0.95\textwidth]{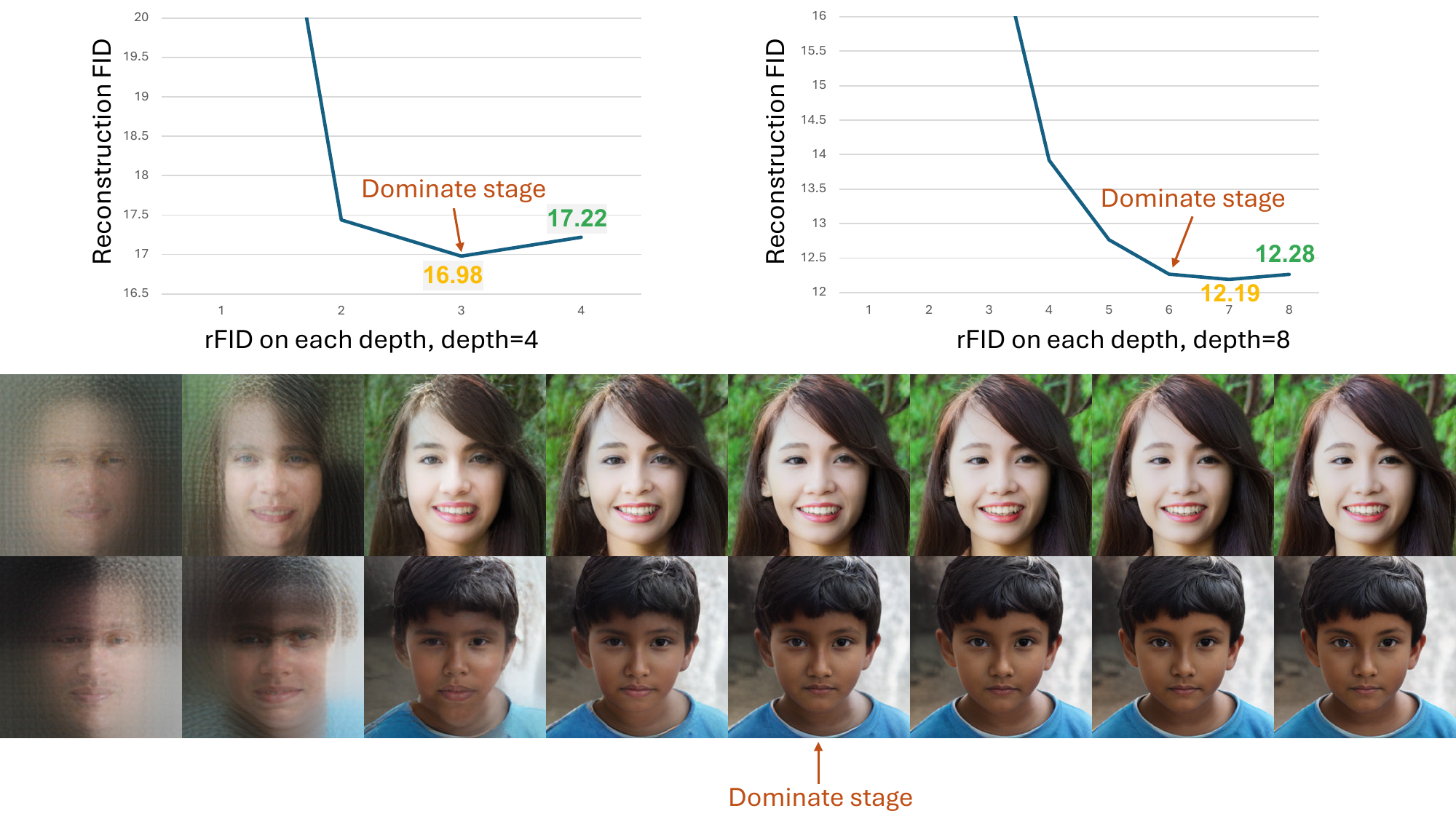}
    \caption{
        Illustration of the invalid encoding issue. The top two figure represent the reconstruction FID of RQVAE on ImageNet with different depths, the bottom figure shows the visualization results on FFHQ. The reconstruction will not continue to improve after the dominate stage.
    }
    \label{fig:invalid-encoding}
\end{figure}

A typical example of variable-length encoding is the RQVAE~\cite{lee2022autoregressive}, which iteratively encodes the reconstruction error within the latent space, aims for achieving increasingly more accurate image reconstructions. However, as illustrated in Figure~\ref{fig:invalid-encoding}, we observed that deeper encoding does not always correlate with better reconstruction quality. We refer to this as the "invalid encoding" issue. To investigate it, we conducted a series of comprehensive experiments, modifying network architectures, learning rates, loss function weights, and codebook sizes, in addition to measuring the frequency of "invalid encoding" across a range of encoding lengths. Table~\ref{tab:invalid} illustrates that the longer the encoding length, the higher probability of encountering the invalid encoding issue. However, we have not ascertained any overarching conclusions regarding the specific conditions that invariably lead to this problem. Below, we offer an intuitive but not rigorous explanation for this issue:

\begin{table}
    \centering
    \begin{tabular}{cccc}
        \toprule
        Depth & 4 & 8 & 16   \\
        \midrule
        Invalid encoding ratio & 0.312 &  0.4 & 0.571   \\
        \bottomrule
    \end{tabular}
    \caption{
        The longer the encoding length, the higher probability of encountering the invalid encoding issue.
    }
    \label{tab:invalid}
\end{table}

Let $D$ represent the decoder and $I$ represent the original input image. The encodings at different depths are denoted by $x_0, x_1, \ldots, x_N$, where N is the depth of encoding, we assume it to 4 in this case. Thus, the reconstruction loss $L$ for the RQVAE can be considered as the combination of the following four reconstruction losses:
\begin{equation}
	\left\{
	\begin{aligned}
		& L_0 = || D(x_0) - I ||_2  \\
		& L_1 = || D(x_0 + x_1) - I ||_2  \\
    & L_2 = || D(x_0 + x_1 + x_2) - I ||_2  \\
    & L_3 = || D(x_0 + x_1 + x_2 + x_3) - I ||_2  
	\end{aligned}
	\right.
\end{equation}

Building on this, we make two assumptions for simplifying the analysis. First, we hypothesize that the decoder functions as a linear transformation, allowing for a more tractable examination of the results. Second, in alignment with standard configurations, we assign equal loss weights to the four losses. With these assumptions, the calculation of the aforementioned reconstruction loss can be simplified as follows:

\begin{equation}
  L_{all} = || D( x_0 + \frac{3}{4} x_1 + \frac{2}{4} x_2 + \frac{1}{4} x_3) - I ||_2
\end{equation}

Therefore, the representation in the latent space that minimizes the image level reconstruction loss would be:

\begin{equation}
   \argmin X = x_0 + \frac{3}{4} x_1 + \frac{2}{4} x_2 + \frac{1}{4} x_3
\end{equation}

This cannot guarantee that  $x_0 + x_1 + x_2 + x_3$ is closer to the $\argmin X$ than $x_0 + x_1 + x_2$. Assuming the encodings at various depths share a common codebook and are independently and identically distributed, the latter sum would definitively be even closer to the ground truth than the former. Therefore, this leads to the invalid encoding issue.

\section*{Question 4: Is diffusion model a maximize likelihood model?}
Autoregressive models are classical maximum likelihood models that facilitate various sophisticated tasks by calculating the likelihood function, including manipulating the generated results and assessing their quality. An interesting question arises: \emph{Can diffusion models also be considered maximum likelihood models?} The initial study on Denoising Diffusion Probabilistic Models~\cite{ho2020denoising} originated from the maximum likelihood to derive the training loss function. ~\cite{song2021maximum} proposes a loss weight setting for ELBO training. Moreover, VDM++~\cite{kingma2024understanding} shows that optimizing with monotonic weightings effectively equates to maximizing the likelihood function with a distribution augmentation. However, in practical training, different loss weights are often employed. The current mainstream practice, exemplified by SD3~\cite{esser2024scaling}, does not fully embrace this principle. 

Coincidentally, a similar perplexity appears in both the generation and evaluation process. During generation, it has been observed that directly sampling from the likelihood model $p_{\theta}(x_{t-1}|x_t,c)$ yields inferior results compared to those modified by classifier-free guidance~\cite{ho2022classifier}: $p_{\theta}(x_{t-1}|x_t) + \lambda (p_{\theta}(x_{t-1}|x_t,c) - p_{\theta}(x_{t-1}|x_t))$, where $\lambda$ is the classifier-free guidance scale. This can be derived~\cite{tang2022improved} as sampling from $p_{\theta}(x_{t-1}|x_t,c)*p_{\theta}(c|x_{t-1})^{\lambda-1}$ . We can easily find that this adjustment merges the likelihood function with the posterior distribution, intimating that maximization of likelihood does not always equate to the best result. This issue is further evidenced during the evaluation phase, wherein models with lower Negative Log Likelihood (NLL) scores do not consistently correspond to the most aesthetically pleasing visual results or the lowest Fréchet Inception Distance (FID) metrics~\cite{ho2020denoising}. This leads to a subtle yet critical question: \emph{Why maximize the likelihood not necessarily lead to optimal results?}

Here is a possible understanding. As illuminated in ~\cite{hyvarinen2005estimation}, score matching is closely related to the maximization of non-normalized likelihood. Typically, score matching can avoid the proclivity in maximum likelihood learning to assign equal probabilities to all data points. In certain special cases, such as Multivariate Gaussian distributions, they are equivariant. VDM++ elucidates that training with monotonic loss weights $w(t)$ indeed equates to maximizing the ELBO for all intermediate states. The particular weighting indicates the varying significance of different noise levels on the ultimate model performance. However, as discussed in Question 2, the image data lacks the "equivariance". In practical training, The \textbf{difficulty} of learning the likelihood function varies with noise intensity; intuitively, the greatest difficulty appears at medium noise levels~\cite{esser2024scaling}, where the likelihood function tends to be less accurately learned. During the generation process, the use of classifier-free guidance can be interpreted as a rectification for the suboptimally learned likelihood function. This is particularly evident in ~\cite{kynkaanniemi2024applying}, the classifier-free guidance is extremely crucial at medium noise levels. During model evaluation, given that tasks at different noise levels have varying degrees of \textbf{importance} to the final result, applying uniform weighting to these NLL losses could not effectively measure the quality of the final generated output.

\section*{Question 5: For diffusion model, how to balance the conflicts among different SNRs?}
As discussed previously, diffusion models differ from autoregressive models in text generation in that they do not maintain "equivariance" across various subtasks. Some studies categorize diffusion models by the noise intensity and explicitly utilize the Mixture of Experts (MOE) strategy for model fitting. Works such as eDiff-I~\cite{balaji2022ediff} and SDXL~\cite{podell2023sdxl} exemplify this approach, with each model do not share parameters. The key of these approaches is the strategic partition of tasks since subtasks not only conflict with each other but also possess relevance. By leveraging these interrelationships, it is possible to improve the efficiency of model convergence and inhibit the exponential increase~\cite{balaji2022ediff} of model parameters.

Other methods attempts to reconcile the conflicts among different noise intensities without increasing the number of parameters. According to VDM++~\cite{kingma2024understanding}, the training objective is the combination of loss weighting and importance sampling:

\begin{equation}
  \mathcal{L}(x) = 
\frac{1}{2} \mathbb{E}_{\epsilon \sim \mathcal{N}(0,\mathbf{I}), \lambda \sim p(\lambda)} 
\left[
\frac{w(\lambda)}{p(\lambda)}
|| \hat{\epsilon}_{\theta}(z_\lambda;\lambda) - \epsilon||_2^2
\right]
\label{eq:training_loss}
\end{equation}

where $\lambda$ denotes the log of Signal-to-Noise Ratios (SNRs), $x$ denotes the training image, $z_\lambda$ is the noisy image at noise intensity $\lambda$. $w(\lambda)$ and $p(\lambda)$ denotes the loss weight and sampling frequency at noise level $\lambda$. 

Therefore, to balance various noise intensities, one might adjust the loss weights or implement importance sampling for different SNRs. MinSNR~\cite{hang2023efficient} is a notable study that carefully designs the loss weights, aiming to circumvent conflicts by pursuing Pareto optimal optimization directions. Studies such as SD3~\cite{esser2024scaling} and HDiT~\cite{crowson2024scalable} have empirically found that increasing the weights for SNRs in the medium range can lead to improved outcomes. As depicted in Equation~\ref{eq:training_loss}, adjusting the loss weights $w(\lambda)$ has similar effects of modifying the sampling frequency $p(\lambda)$. However, in practice, increasing the loss weight for important tasks is equivariant to raising the learning rate, while enhancing the frequency can be seen as dedicating more computational resources (Flops), which often results in better performance.

From another perspective, performing importance sampling on different noise levels can be considered as designing a noise schedule~\cite{hang2024improved}, or a type of signal decomposition, as we discussed in Question 2. When applying independent and identically distributed (i.i.d.) Gaussian noise across different spatial positions, previous studies have indicated the necessity to adapt the noise schedule based on token length~\cite{chen2023importance}, and ensuring there is no signal leakage in the final step~\cite{lin2024common,tang2023volumediffusion}. ~\cite{hang2024improved} empirically introduced a noise schedule for training that achieves more efficient convergence and has proven its effectiveness across diverse conditions. However, it may need to adjust hyperparameters based on the target distribution, and lacks analysis of the conflicts during inference. The author speculate that abandoning the use of i.i.d. Gaussian noise for signal decomposition might be a fundamental approach to resolving the conflicts.

\section*{Question 6: Is there a scaling law for diffusion models?}
Upon rethinking the remarkable achievements of Large Language Models (LLMs), one pivotal factor is the principle of the scaling law. This naturally raises a question: \emph{Do diffusion models in visual generation also conform to a scaling law?}

The challenge in addressing this question lies in the lack of an evaluation metric that aligns with human perception to assessing the performance of the model. In language modeling through autoregressive models, the principle of "equivariance" facilitates the evaluation of model performance by assigning equal weight to the negative log likelihood loss across different subtasks during both training and evaluation phases. In contrast, the subtasks within diffusion models lack this equivariance; they each contribute to the final generated output in varying degrees. Therefore, a simple, equal-weighted aggregation of losses fails to capture the full extent of a generative model's capabilities. 

\begin{figure}
    \centering
    \includegraphics[width=0.95\textwidth]{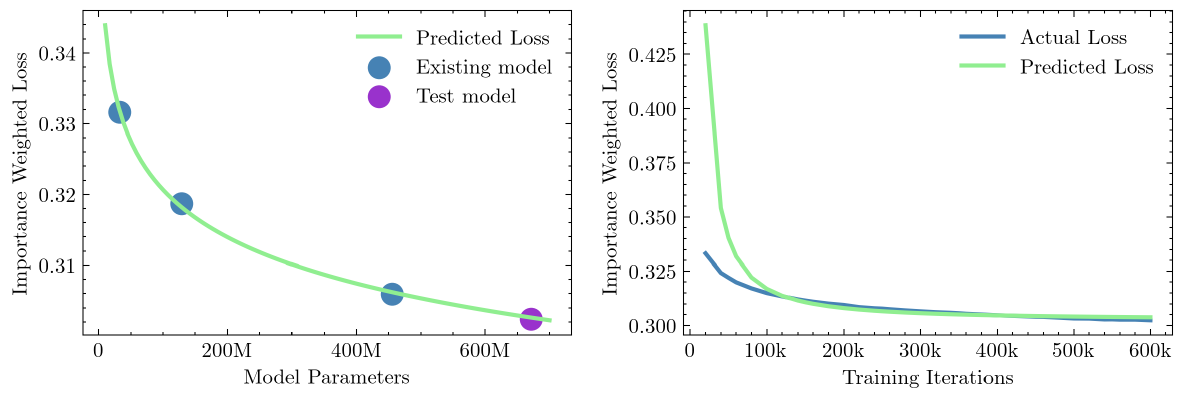}
    \caption{
        The diffusion scaling law rely on importance weighted loss. For the left figure, $y=0.369*x^{-0.030}$. For the right figure, $y=0.303+1.4*10^5*x^{-0.140}$.
    }
    \label{fig:ScalingLaw}
\end{figure}

To uncovering an appropriate metric, the first attempt is to construct an importance coefficient for various subtasks. The novel noise schedule defined in ~\cite{hang2024improved} can be considered as assigning "difficulty coefficients" to distinct tasks. We treat it as an "importance coefficient", and employing it to weight the losses at different noise intensities. For training text2image models, we employed COYO~\cite{kakaobrain2022coyo-700m} dataset, which consists of 700M text-image pairs. We trained four models with parameter counts of 32.28M, 128.56M, 454.98M, and 671.32M respectively. For convenience, we denote them as S, M, L, and XL. All the models are trained with 1024 batchsize. We use the "importance weighted loss" as metric to measure the model's performance. Following the approach outlined in ~\cite{su2024unraveling}, we utilize the formula from ~\cite{kaplan2020scaling} to estimate the performance of the XL model based on the S, M, L models. The results are shown in Figure~\ref{fig:ScalingLaw}. The left figure predicts the model's performance according to the number of parameters, while the right figure predicts the model's performance in relation to the number of training iterations. It can be seen that both predictions are quite accurate, although the scale of verification is relatively limited due to resource constraints. Nevertheless, it's essential to note that, there is no direct evidence to affirm that the metric aligns with human judgment.

The second strategy employs established metrics for evaluating generative models, with Fréchet Inception Distance (FID)~\cite{heusel2017gans} being the most prevalent. FID is designed to quantify the disparity between two distributions of data. However, when dealing with large-scale generative models and exceedingly complex data distributions, it becomes challenging to accurately capture the target distribution with a finite sample set, inevitably leading to biases in the FID score. Furthermore, FID assumes that the feature vectors extracted from the neural network follows Gaussian distribution, which introduces significant systematic errors. It is also important to note that some studies~\cite{parmar2022aliased,gu2020giqa,borji2019pros} have highlighted further issues associated with the FID metric.

To ensure alignment with human preferences, the most crucial approach involves using extensive human annotation as the benchmark for evaluation. Taking text2image generation as an example, one potentially approach is to collect a large number of high-quality text-image pairs. For the generative models to be tested, they can generate results based on the given text prompts. Users evaluate which image, the generated one or the ground truth image, better satisfies their preferences. In theory, as the quality of the models improves, this preference rate of generated results should converge towards 0.5. This preference rate could then act as an indicator for the scaling law, providing insights into how computational resources, model size, and data size affect the model's final performance. It is important to note that the quality of outputs from diffusion-based vision models is highly sensitive to the chosen inference strategy~\cite{karras2022elucidating,ho2022classifier}, which differs significantly from large language models. With a metric that can capture the human preference, this factor merits further exploration.

\section*{Acknowledgements}
I would like to thank Tiankai Hang for the consistent discussions during this work. I would also want to thank all the friends who have discussed these issues with me, including Dong Chen, Zhicong Tang, Zheng Zhang, Jianmin Bao, and Chengyi Wang. Moreover, we sincerely hope that readers will point out any errors or inadequacies in our understanding, or feel free to directly contact me, to collectively strive to resolve these challenges in the field of visual generation.

Finally, if you plan to cite this paper and use an abbreviation, hope you can use VSD or Visual Signal Decomposition. The author would greatly appreciate it.

\clearpage
{\small
\bibliographystyle{plainnat}
\bibliography{egbib}
}

\end{document}